\documentclass{stl}

\usepackage{latexsym}
\usepackage{amssymb}
\usepackage{amsmath}
\usepackage{amsthm}
\usepackage{booktabs}
\usepackage{enumitem}
\usepackage{graphicx}
\usepackage{dsfont}
\usepackage{color}
\usepackage{url}
\DeclareMathOperator*{\argmax}{argmax}

\newcommand{\BibTeX}{B\kern-.05em{\sc i\kern-.025em b}\kern-.08em\TeX}

\begin{document}


\begin{frontmatter}


\paperid{123} 


\title{Cross-Camera Data Association via\\
GNN for Supervised Graph Clustering}


\author{\fnms{{\DJ}or{\dj}e}~\snm{Nedeljkovi{\'c}}}

\address{Independent Researcher}


\begin{abstract}
Cross-camera data association is one of the cornerstones of the multi-camera
computer vision field. Although often integrated into detection and tracking tasks
through architecture design and loss definition, it is also recognized as an
independent challenge.
The ultimate goal is to connect appearances of one item from all cameras,
wherever it is visible. Therefore, one possible perspective on this
task involves supervised clustering of the affinity graph, where nodes are
instances captured by all cameras. They are represented by appropriate visual
features and positional attributes.
We leverage the advantages of GNN (Graph Neural Network) architecture to
examine nodes' relations and generate representative edge embeddings. These
embeddings are then classified to determine the existence or non-existence of
connections in node pairs. Therefore, the core of this approach is
graph connectivity prediction.
Experimental validation was conducted on multicamera pedestrian
datasets across diverse environments such as the laboratory, basketball court,
and terrace. Our proposed method, named SGC-CCA, outperformed the state-of-the-art
method named GNN-CCA across all clustering metrics,
offering an end-to-end clustering solution without the need for graph
post-processing. The code is available at \url{https://github.com/djordjened92/cca-gnnclust}.
\end{abstract}

\end{frontmatter}


\section{Introduction}

The latest works on multi-camera tasks (multi-target tracking, 3D object
detection, pedestrian pose estimation, etc.)
\citep{huang2022bevdet4d, huang2022bevdet, liu2022petr, liu2023petrv2, wang2021detr3d}
usually employ the end-to-end methods, leveraging the very popular transformers architecture. They often
provide detections from all camera views to the transformer decoder and
rely on the neural network capability to learn instance associations
implicitly. However, some methods contain cross-camera instance association
as an explicit step in the solution pipeline. This component usually makes
a difference, so we find it worth studying as an independent topic.

The most common perspective of this problem in literature is decoupling
the task into bipartite matching of instances for a particular pair of
cameras. The next step is usually the aggregation of camera-pairs results in
a global solution. Hungarian algorithm, which is used often in bipartite
graph matching exploits association matrix consisting of instance distances.
This distance usually reflects the difference in ReIdentification score
(obtained from some appearance embedding model) and 3D/2D spatial distance
of instances. This approach suffers from high computational demands
as the number of instances grows. It also relies on manual thresholding
when it comes to association decisions. Moreover, when pairs of views are
analyzed independently it can lead to inconsistency when aggregation
of the results is performed.

One of the recent works is an attempt to find a global solution for
instance association, incorporating all camera views simultaneously.
Named GNN-CCA \citep{luna2022graph}, this method exploits GNN Message Passing
architecture on top of the complete graph connecting all instances across views.
The use case, which we also used in this work, involves associating pedestrians
captured by four cameras in three different environments. Images are sampled
from video sequences, with pedestrian annotations provided as ground-truth bounding boxes.
Some of the best person ReIdentification models provided
appearance embeddings for each image crop. Based on the available camera
calibration parameters and annotations, the authors managed to provide
the ground plane coordinates of each person for each camera view as well.
Both appearance and position are used to find a good representation
of each node and edge. Following several iterations of message passing, the GNN-CCA \citep{luna2022graph}
method employs binary classification on each edge. This classifier should
determine if two connected nodes belong to the same person or not.
The final and key step of this method is graph post-processing,
consisting of two operations: \textit{edge pruning} and \textit{graph splitting}.
These heuristic techniques aim to yield the final set of node clusters
(connected components) that represent all visible person identities.

\paragraph{Contributions}
This work provides a novel perspective of an instance matching problem as an
end-to-end supervised graph clustering task, unlike the GNN-CCA \citep{luna2022graph} approach,
in which this problem is reduced to the linkage prediction of nodes.
Consequently, the following contribution is the avoidance of any tailored graph
post-processing heuristics necessary for GNN-CCA \citep{luna2022graph} approach.
Standard graph clustering inference assumes that one large graph is provided as
an input that requires partitioning. The IMDB-test, one of the test sets used in the
Hi-LANDER \citep{xing2021learning} paper, consists of 50289 entities/clusters with an average cluster
size of 25 (number of cameras in our case). We have shown that the same GNN model can
be effectively trained and tested on many smaller, independent input graphs
(one graph per scene), with careful adaptations, as presented in Section 2.
Furthermore, the proposed SGC-CCA method surpasses GNN-CCA \citep{luna2022graph}
in all clustering metrics.

\begin{figure*}[h]
\centering
\includegraphics[scale=1.0]{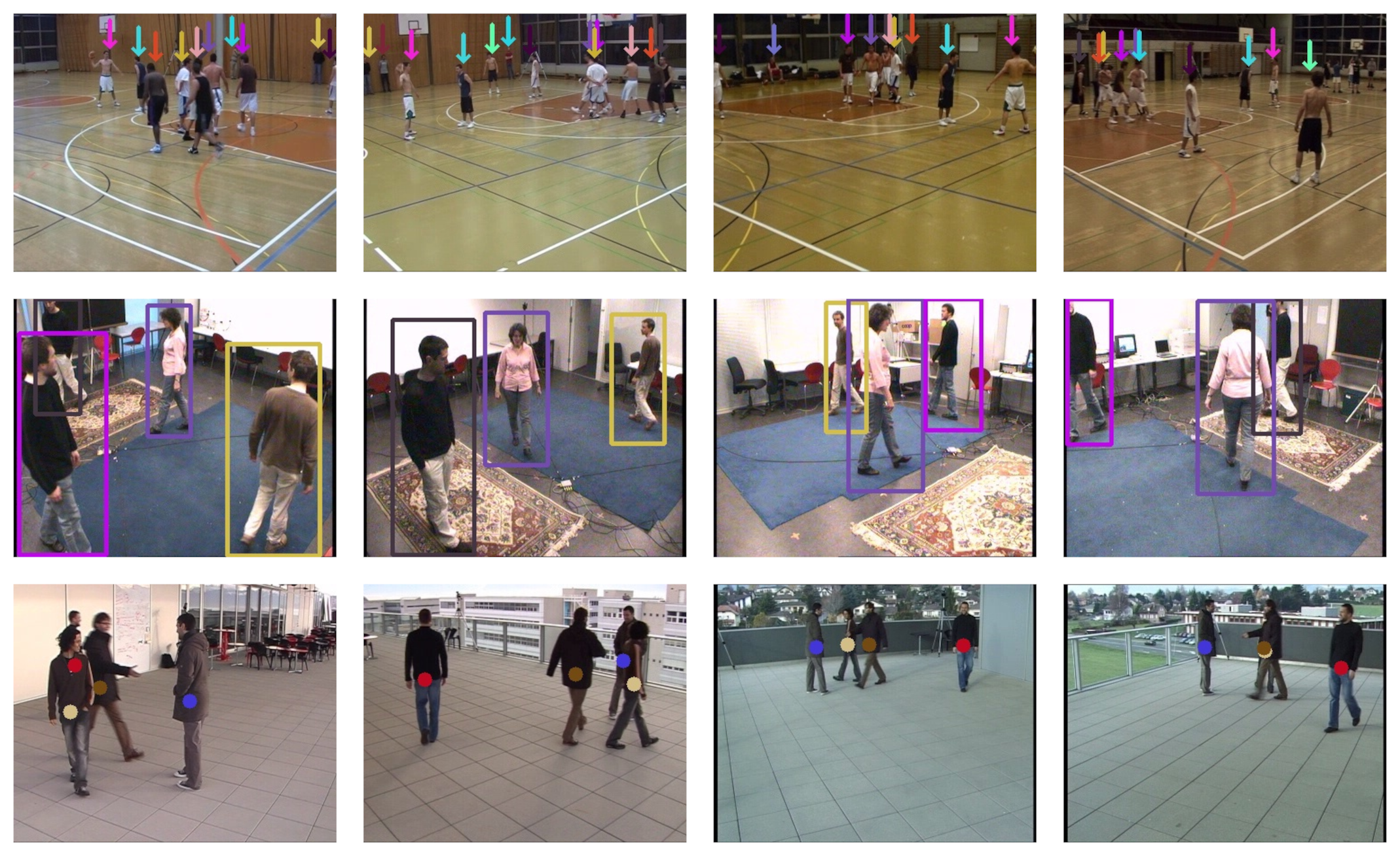}
\vspace{-10pt}
\caption{Sample scenes from EPFL sequences: Basketball, Laboratory and Terrace.}
\vspace{5pt}
\label{fig:seqs_preview}
\end{figure*}

\subsection{Why Hi-LANDER?}

The motivation for learnable clustering methods via GNN lies in the aim to overcome
some limitations of the traditional techniques. K-Means \citep{lloyd1982least}
requires the number of clusters as an input and assumes that clusters are convex-shaped.
Spectral clustering \citep{shi2000normalized} demands clusters balanced in the
number of instances. The shape-agnostic technique is DBSCAN \citep{ester1996density},
which introduced a density-based approach. It brought also a minimal requirement of
domain knowledge and good efficiency on the large database at the time. However,
it assumes that different clusters have similar densities. The family of linkage-based
clustering methods overcomes all mentioned dependencies on the mentioned assumptions.
The well-established method of this kind is Rank-Order Clustering \citep{zhu2011rank}
and its more efficient successor Approximate Rank-Order Clustering (ARO)
\citep{otto2017clustering}. Although these techniques proposed new cluster-level affinity
(distance) with demonstrated ability to detect noise instances, instead of traditional
$l_{1}$/$l_{2}$ distance, they are still based on the linkage likelihood estimation
by different heuristics.

To mitigate the issues of using arbitrary thresholds, \citet{wang2019linkage} introduce
a novel approach employing a learnable clustering framework leveraging the graph
convolution network (GCN)\citep{kipf2017semisupervised}.
At the heart of their method lies the creation of what they term the Instance Pivot
Subgraph (IPS) around each instance acting as a pivot, determined by nearest
neighbour criteria. Within each IPS, connectivity between the pivot and its neighbors
is assessed using GCN as a learnable component. This assessment is based on node
classification, distinguishing between positive (pivot-like) and negative nodes.
Only positive nodes within each IPS are connected to the pivot. Following
connectivity prediction across all IPSes, the authors employ link merging to
derive the final set of clusters.

Another example of the supervised clustering \citep{yang2019learning} also leverages
GCN \citep{kipf2017semisupervised} to tackle the challenge posed by variations of the cluster
patterns. The method starts from the kNN affinity graph which encompasses the whole
dataset. The first stage is generating Cluster Proposals which is the heuristic-based
approach of subgraphs selection. Those size-varying subgraphs represent multi-scale
candidates for the resulting cluster partitioning. The second stage is
Cluster Detection where the GCN-D model learns to detect which proposals are kept as good
enough and which ones are dropped. The final stage is cluster refinement of the top proposals
identified by GCN-D, utilizing another (GCN-S) model. This step is called Cluster
Segmentation. Unlike graph-level prediction in the Detection stage, this model outputs
a probability for each node to show if it is a valid cluster member or an outlier.
The described method relies on the numerous generated subgraphs. The same
research group aimed to overcome this redundant computation by employing a fully learnable
clustering paradigm in \citep{yang2020learning}. This method is based on the vertex
confidence estimator (GCN-V) - indicates if the vertex belongs to the specific class and
edge connectivity estimator (GCN-E) - outputs the probability that specific edge connects
two incident vertices. After the affinity graph processing by these two learnable
components, it is possible to create a directed path over detected edges from vertices
with lower confidence to those with higher confidence. This process deduces several
isolated trees, partitioning the affinity graph into clusters.

The Hi-LANDER \citep{xing2021learning} belongs to the same group of supervised GNN-based methods,
focusing on linkage prediction as the foundation of clustering. In contrast to
earlier single-partitioning techniques \citep{wang2019linkage, yang2019learning, yang2020learning},
Hi-LANDER \citep{xing2021learning} stands out as the first hierarchical, agglomerative
clustering method employing GNN architecture.
This multi-level algorithm allows for modeling the "natural granularity"
of the data. Unlike previous approaches, Hi-LANDER \citep{xing2021learning} employs
\emph{full-graph} inference and predicts connectivity based on \emph{edge} features.
Unlike \citep{yang2020learning}, Hi-LANDER \citep{xing2021learning} utilizes a single
model for both linkage probability estimation and node density prediction.
Node density serves regularization purposes and facilitates additional edge
refinement. This method demonstrates superior efficiency alongside significant
improvements in metrics. Further details about this approach will be discussed
in the following section.

\vspace{0pt}
\begin{table*}[h]
\caption[]{Clustering pipeline in two levels (as table rows), based on the peak estimation steps (as table columns)}
\vspace{10pt}
\centering
\begin{tabular}{c|c|c|c|}
\hline
\boldmath{$l_1$}
&
\includegraphics[scale=0.55]{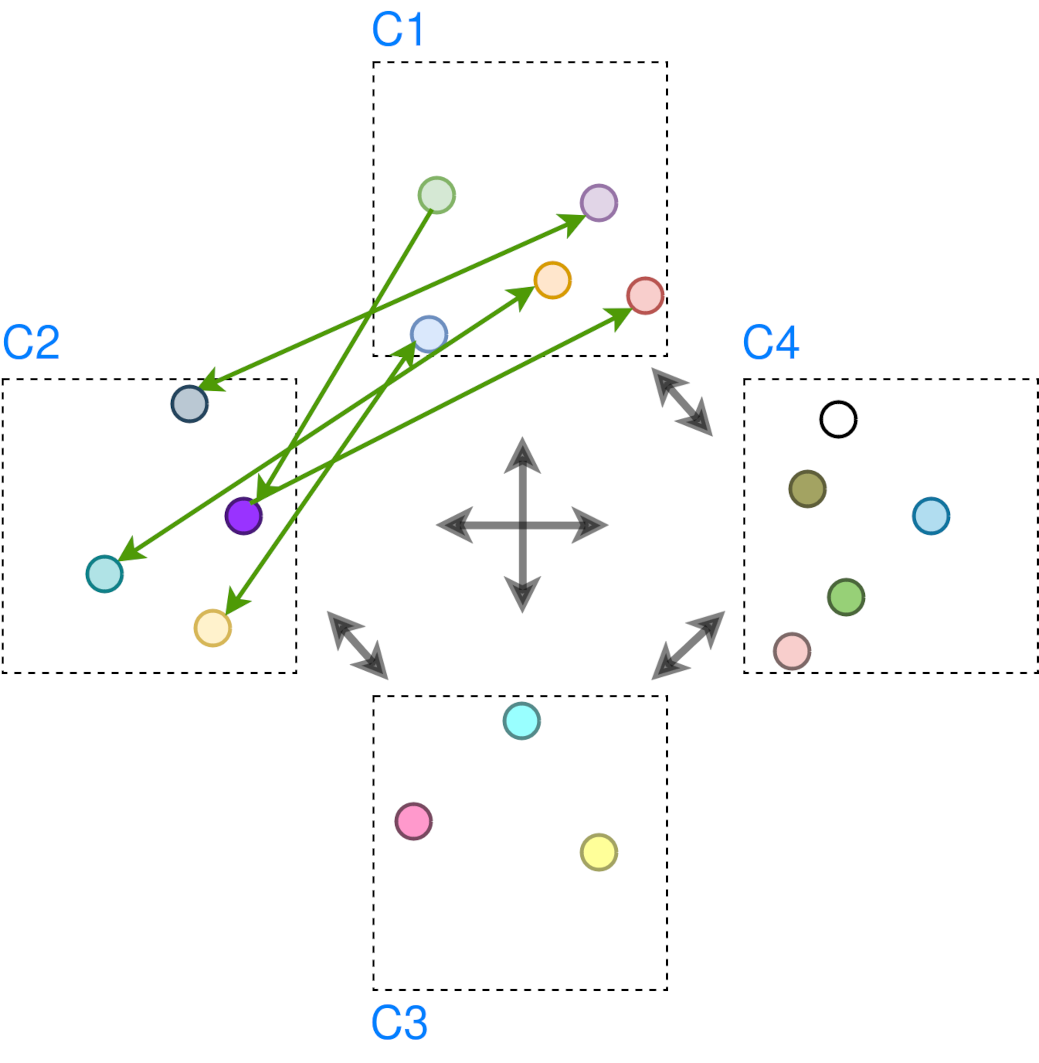}
&
\includegraphics[scale=0.55]{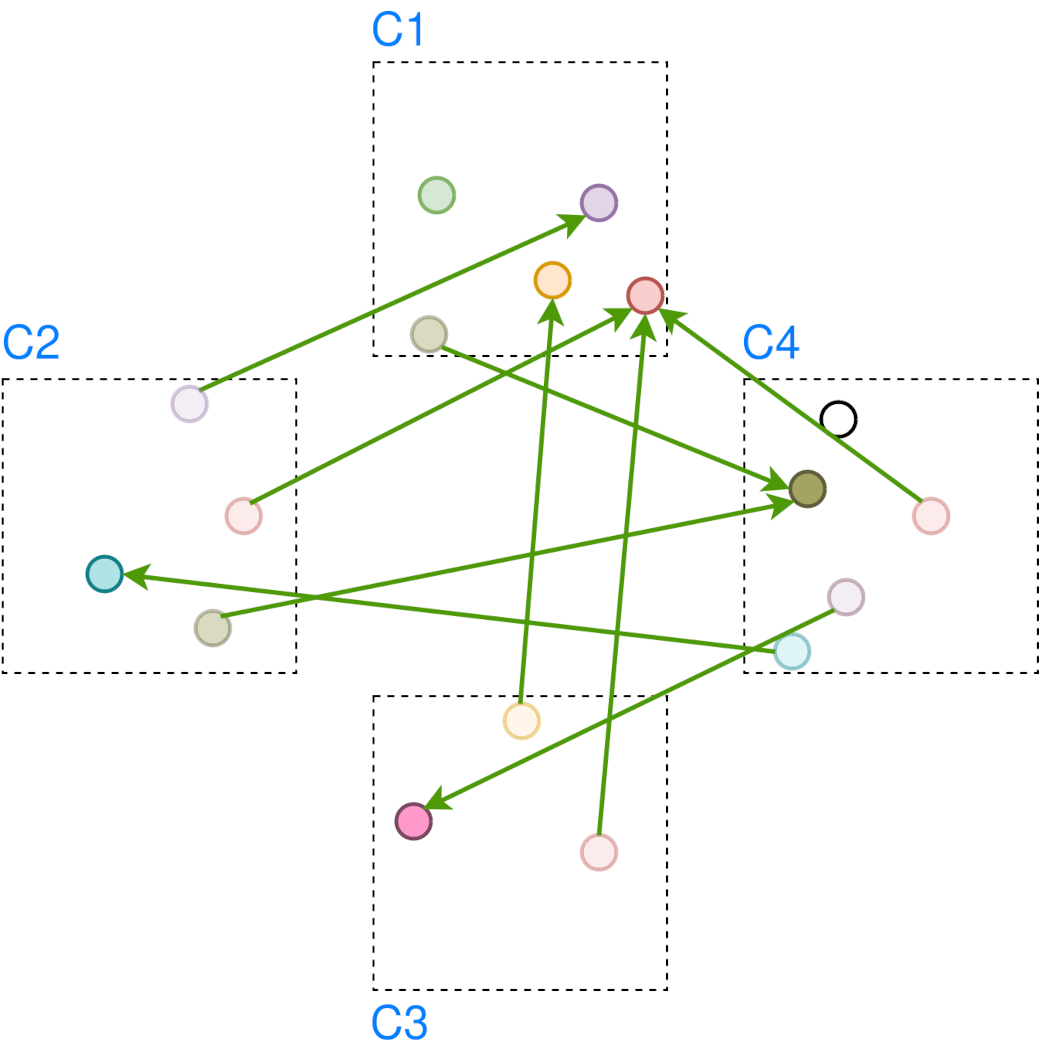}
&
\includegraphics[scale=0.55]{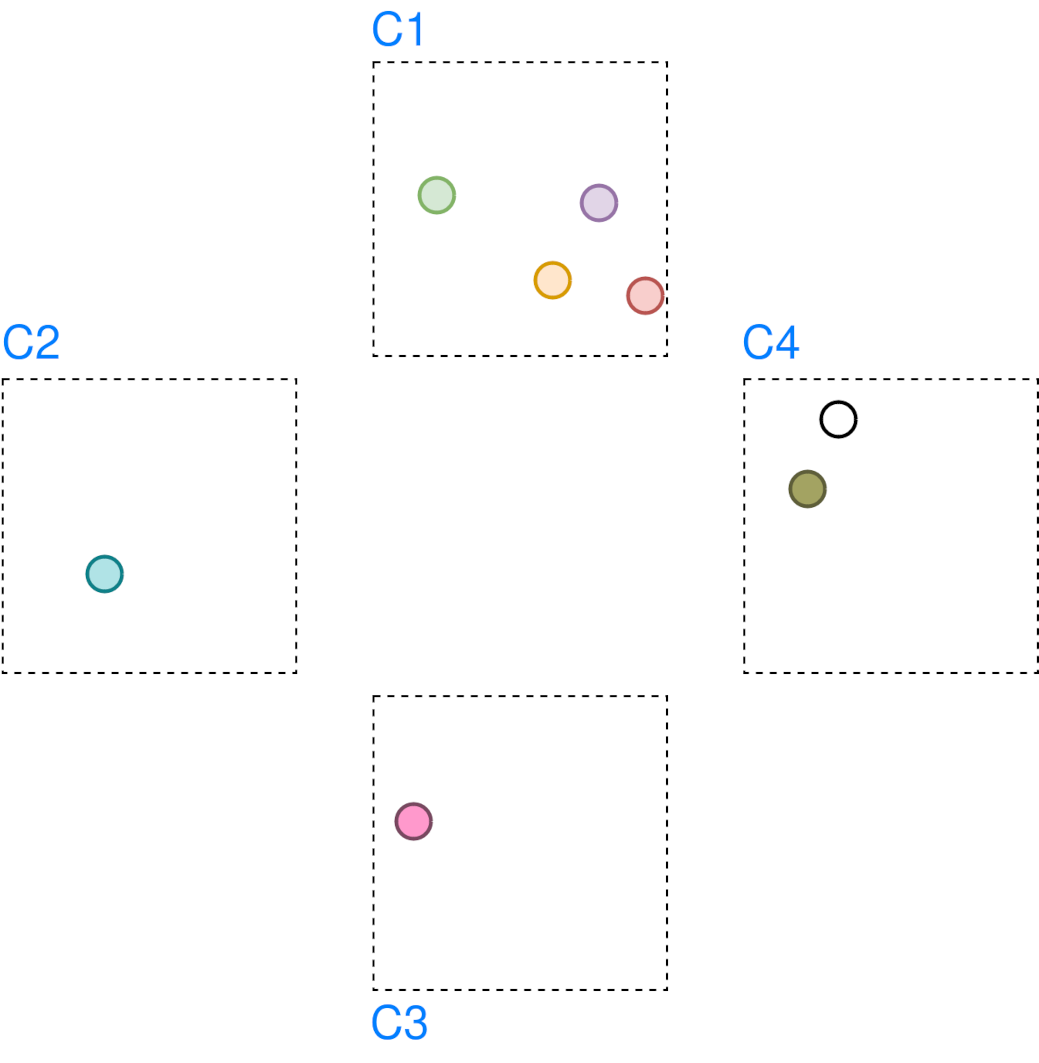}
\\
\hline
\boldmath{$l_2$}
&
\includegraphics[scale=0.55]{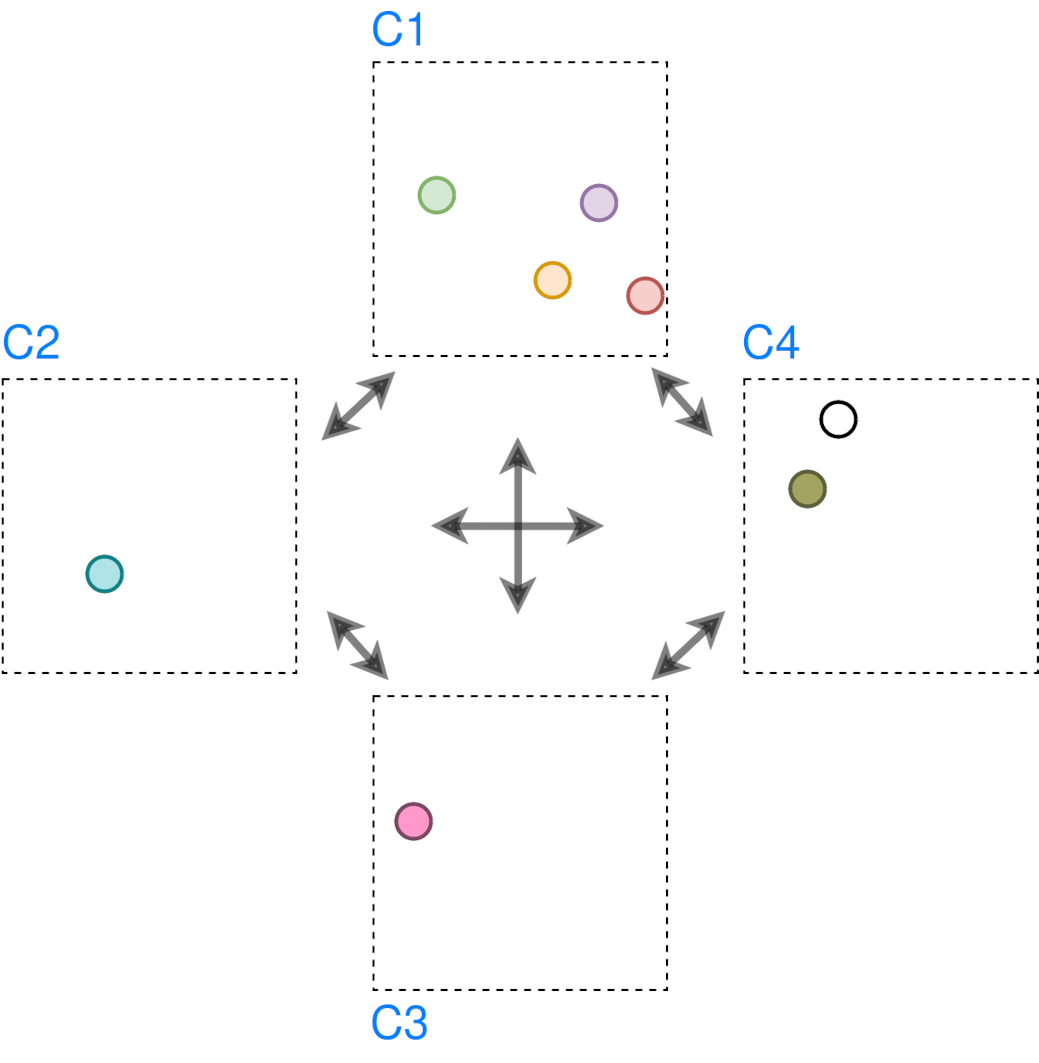}
&
\includegraphics[scale=0.55]{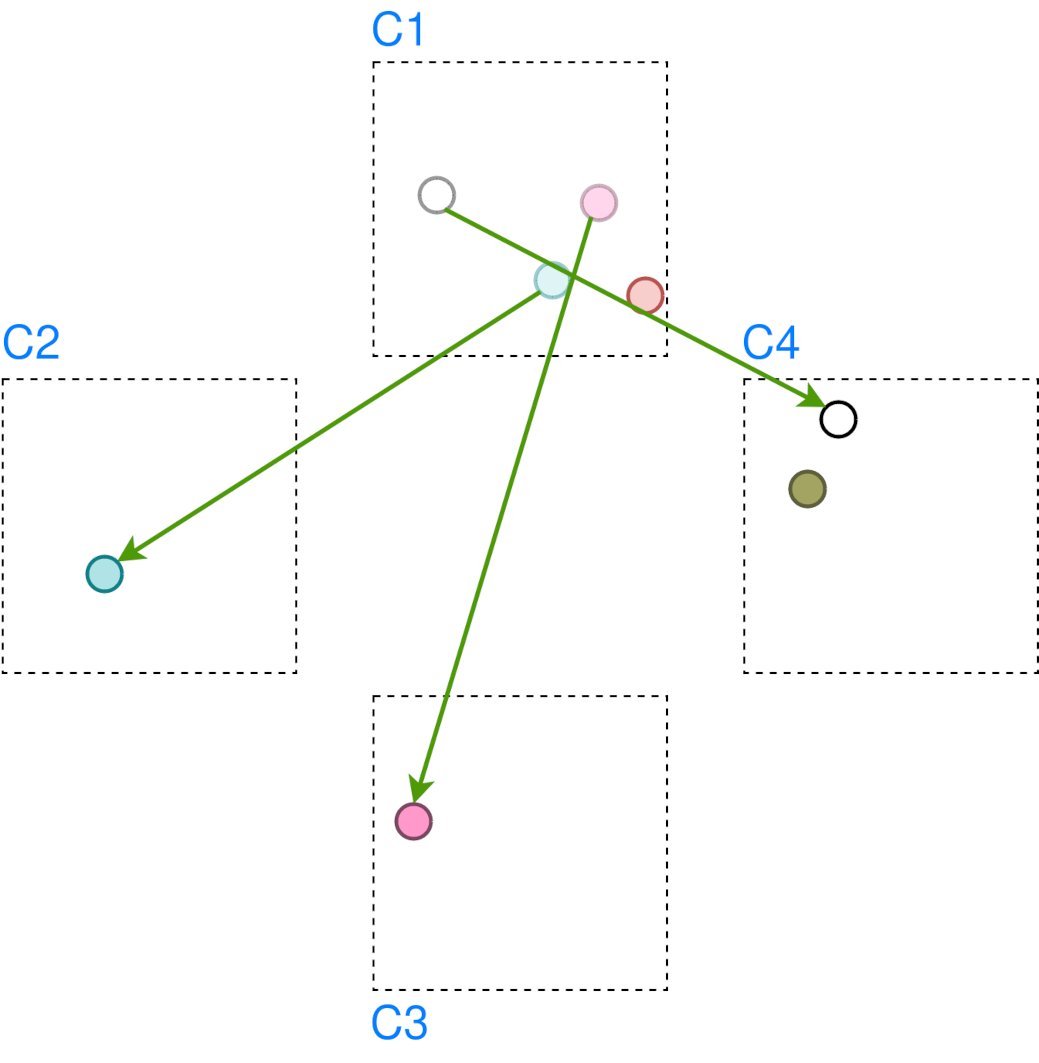}
&
\includegraphics[scale=0.55]{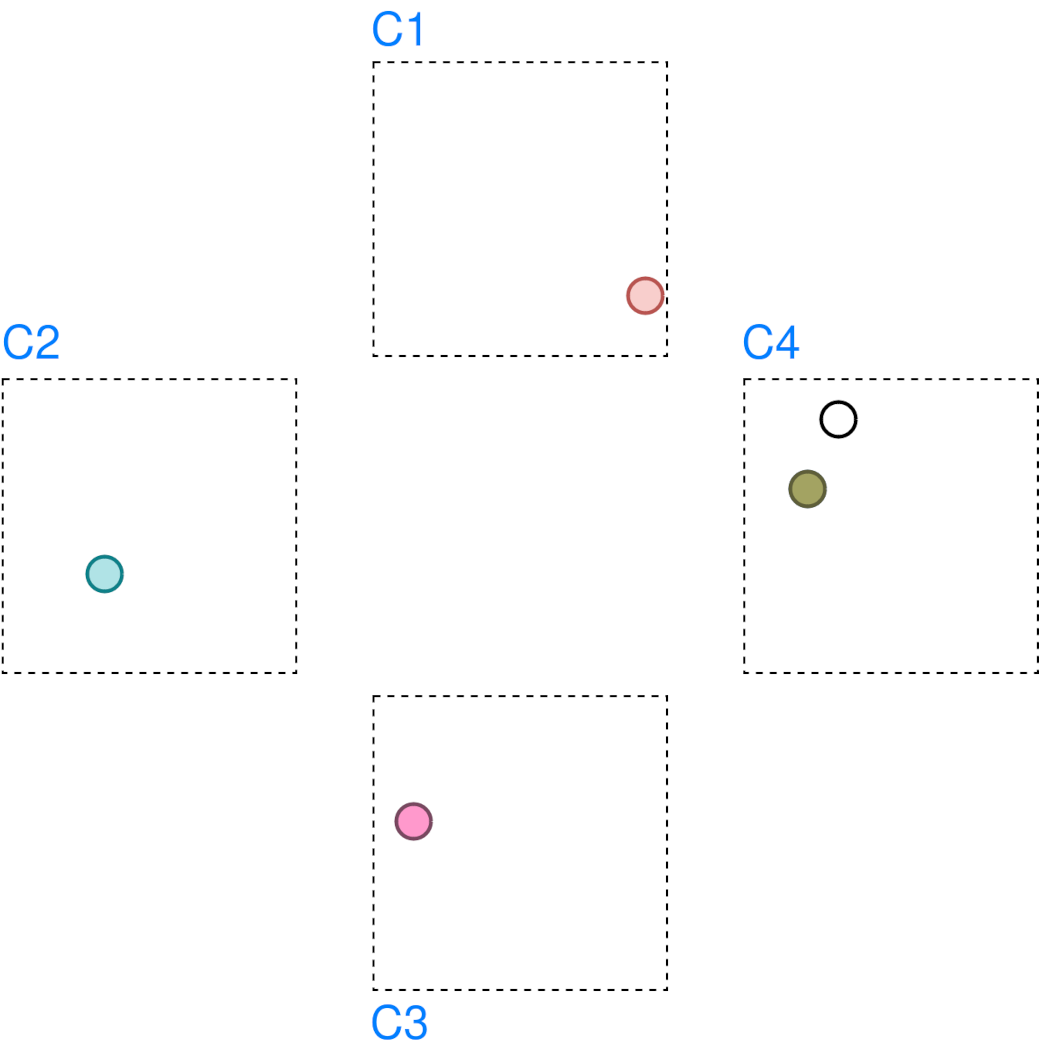}
\\
\hline
&\textbf{a) affinity graph} & \textbf{b) graph decoding} & \textbf{c) peaks}
\\
\hline
\end{tabular}
\label{tb:pipeline}
\vspace{10pt}
\end{table*}

\section{Methodology}

This work applies supervised graph clustering to perform cross-camera instance
association. We adapt Hi-LANDER \citep{xing2021learning} hierarchical GNN
architecture to predict graph connectivity and to extract connected components
of graph which successfully represent person identities. Although this technique
is applicable to the arbitary number of views and types of objects, we focused
on the EPFL\footnote{https://www.epfl.ch/labs/cvlab/data/data-pom-index-php} pedestrian video sequences with four views. Thus, our results are
comparable with GNN-CCA \citep{luna2022graph} results. Sample scenes of all
three setups are shown in Figure~\ref{fig:seqs_preview}. The same person across
all views is marked with the same color.

The following sections present the implemented clustering and training pipeline
from the Hi-LANDER \citep{xing2021learning}. Changes that are part of the SGC-CCA
customization will be \textbf{emphasized}.

\subsection{Graph Creation}

Let $M$ be the number of cameras in the setup. $N$ is the total number of pedestrian
bounding boxes across all views at the timestamp $t$. The set of cameras is denoted as
$C = \{c_i \mid i \in [1, M]\}$ and $[r_1, r_2, ..., r_N]$ are bounding box labels
from the set $R = \{r_i \mid i \in [1, O]\}$ of $O$ pedestrian identities.
$F = \{f_i \mid i \in [1, N]\}$ is the set of visual features extracted
on pedestrian crops using a pretrained ReID model.

The input structure for this method is a directed graph $G = (V,E)$, where
$V = \{v_i \mid i \in [1, N]\}$ represents the set of nodes denoting all pedestrian
bounding boxes. Each node is depicted by embedding $h_i$ initialized with
appropriate, normalized feature $f_i$, forming node embeddings set $H = \{h_i \mid i \in [1, N]\}$.

The pedestrian $v_i$ on the camera $c_j$ has a ground-truth bounding box in the form
$(x_i, y_i, w_i, h_i)$ - upper-left corner coordinates, width, and height. Estimation
of the pedestrian's standing point is the middle point of bounding box lower edge
$(x_i + \frac{w_i}{2}, y_i)$. Thanks to the available camera's intrinsic and extrinsic
parameters, it is possible to create a homography matrix $H_{c_j}$. It enables projection
from the camera $c_j$ plane into the common ground plane. Thus, the standing point position in
the common ground plane $(\hat{x_i}, \hat{y_i})$ is calculated with the:
\begin{eqnarray}\label{eq:homography}
\begin{bmatrix}
\hat{x}_i\\
\hat{y}_i\\
1
\end{bmatrix} = H_{c_j}\begin{bmatrix}
    x_i + \frac{w_i}{2}\\
    y_i\\
    1
\end{bmatrix}
\end{eqnarray}

\textbf{The node distance combines the cosine distance of node embeddings and normalized Euclidean
distance of positions $(\hat{x}, \hat{y})$}, comparing to the original work where the cosine distance
is used only. Therefore the distance between nodes
$v_i$ and $v_j$ is defined as:
\begin{eqnarray}\label{eq:distance}
m_{ij} = (1 - \langle h_i, h_j \rangle)
\frac{{\lVert (\hat{x}_i - \hat{x}_j), (\hat{y}_i - \hat{y}_j) \rVert}_2}
{\underset{p, q}{max}{\lVert (\hat{x}_p - \hat{x}_q), (\hat{y}_p - \hat{y}_q) \rVert}_2}
\end{eqnarray}
where $\langle \cdot \; , \; \cdot \rangle$ denotes the inner product of two vectors, which
is equal to cosine similarity for normalized embedding vectors.

For each node of camera $c_i$, we find the \textbf{one closest neighbor from each other camera
view} $c_j, j \neq i$ (cell $(l_1, a)$ in Table~\ref{tb:pipeline}), unlike Hi-LANDER \citep{xing2021learning}
which applies pure kNN over the whole corpus of nodes. This neighbor selection per camera is
related to the setup where the pedestrian can appear mostly once in each view. Consequently,
each node adds $M - 1$ edges to the set of graph edges $E$. If the node $v_j$ is the neighbor
of the node $v_i$, adjacency matrix score $A(i, j)$ is assigned a cosine similarity of
node embeddings $\langle h_i, h_j \rangle$.

\subsection{Graph Clustering} \label{clustering}

\subsubsection{Graph Encoding} \label{encoding}
The input affinity graph is processed by graph convolution network (GCN)\citep{kipf2017semisupervised}.
This message passing paradigm \citep{gilmer2017neural} simulates nodes' interaction
and information exchange. Using $h_i$ as the input embedding of the node $v_i$, GCN
encodes it as a new node embedding $h_i'$ in the following way:
\begin{eqnarray}\label{eq:gcn}
h_i' = \phi(h_i, \sum_{v_j \in N_{v_i}} w_{ji}\psi(h_j))
\end{eqnarray}
where $\phi$ and $\psi$ are MLPs, $w_ji$ is a trainable vector. $N_{v_i} = \{v_j, (v_j, v_i) \in E\}$
is the neighborhood of node $v_i$, defined with the set of incoming edges.

GCN encoder can be applied multiple times on the same graph, so the effect of the number of
message passing steps is also explored in this work.

\subsubsection{Linkage Prediction and Node Density} \label{linkage prediction}
After the Graph Encoding step, resulting node features $H'$ are used to predict the linkage
between nodes. The edge $(v_i, v_j)$ connectivity is predicted by applying MLP classifier
$\theta$ from Equation~(\ref{eq:edge_cls}). The input is a vector created from \textbf{concatenated node
features} ($h_i', h_j'$) and \textbf{nodes' ground plane positions}
$(\hat{x_i}, \hat{y_i})$, $(\hat{x_j}, \hat{y_j})$.
The original work considers the concatenation of node features only.
The output is a sigmoid activation which estimates the probability that two connected nodes
have the same label.
\begin{eqnarray}\label{eq:edge_cls}
\hat{r}_{ij} = P(r_i = r_j) = \sigma(\theta([h_i', \hat{x_i}, \hat{y_i}, h_j', \hat{x_j}, \hat{y_j}]^T))
\end{eqnarray}

A node density $d_i$ is the value that depicts the weighted partition of neighbors which have the
same label as the node $v_i$. Its estimation is defined as:
\begin{eqnarray}\label{eq:d_est}
\hat{d_i} = \frac{1}{k}\sum_{j=1}^{k}\hat{e}_{ij}a_{ij}.
\end{eqnarray}
where $a_{i,j} = \langle h_i, h_j \rangle$ is the similarity of nodes' embeddings, and
$\hat{e}_{ij}$ is the edge coefficient defined as:
\begin{eqnarray}\label{eq:edge_coeff}
\hat{e}_{ij} = P(r_i = r_j) - P(r_i \neq r_j).
\end{eqnarray}

\subsubsection{Graph Decoding} \label{decoding}
After an estimation of the graph attributes (node density and edge coefficient) using the GNN encoder,
it is possible to find connected components of the graph in the next two steps:
\begin{enumerate}
    \item \textit{Edge filtering}: We initialize a new edge set $E' = \emptyset$. The subset of
    outgoing edges for each node $v_i$ are created as
    \begin{eqnarray}\label{eq:edge_subset}
    \varepsilon(i) = \{j \mid (v_i, v_j) \in E \wedge \hat{d}_i \leq \hat{d}_j \wedge \hat{r}_{ij} \geq p_{\tau}\}
    \end{eqnarray}
    where $\hat{r}_{ij} = P(r_i = r_j)$ and $p_{\tau}$ is the edge connection threshold. Each node
    with non-empty $\varepsilon_i$ contributes to the set $E'$ with one edge selected as
    \begin{eqnarray}\label{eq:edge_filter}
    j = \argmax_{k \in \varepsilon(i)} \hat{e}_{ik}
    \end{eqnarray}
    The edge $(v_i, v_j)$ is added to the $E'$. With the condition $\hat{d}_i \leq \hat{d}_j$
    authors introduced an inductive bias to discourage connection to nodes on the border of
    clusters.
    \item \textit{Peak nodes}: The set of edges $E'$ defines new, refined graph $G'$
    (cell $(l_1, b)$ in Table~\ref{tb:pipeline}) on the same set of nodes. The peak nodes are those without outgoing edges.
    They have a maximum density in the neighborhood. The way $G'$ is created implies
    a separation of the graph in the set of connected components $Q = \{q_i \mid i \in [1, Z]\}$.
    Consequently, each connected component has one peak node distinguished by the highest
    density in the connected component (cell $(l_1, c)$ in Table~\ref{tb:pipeline}).
\end{enumerate}

\begin{figure}[h]
    \centering
    \includegraphics[scale=0.8]{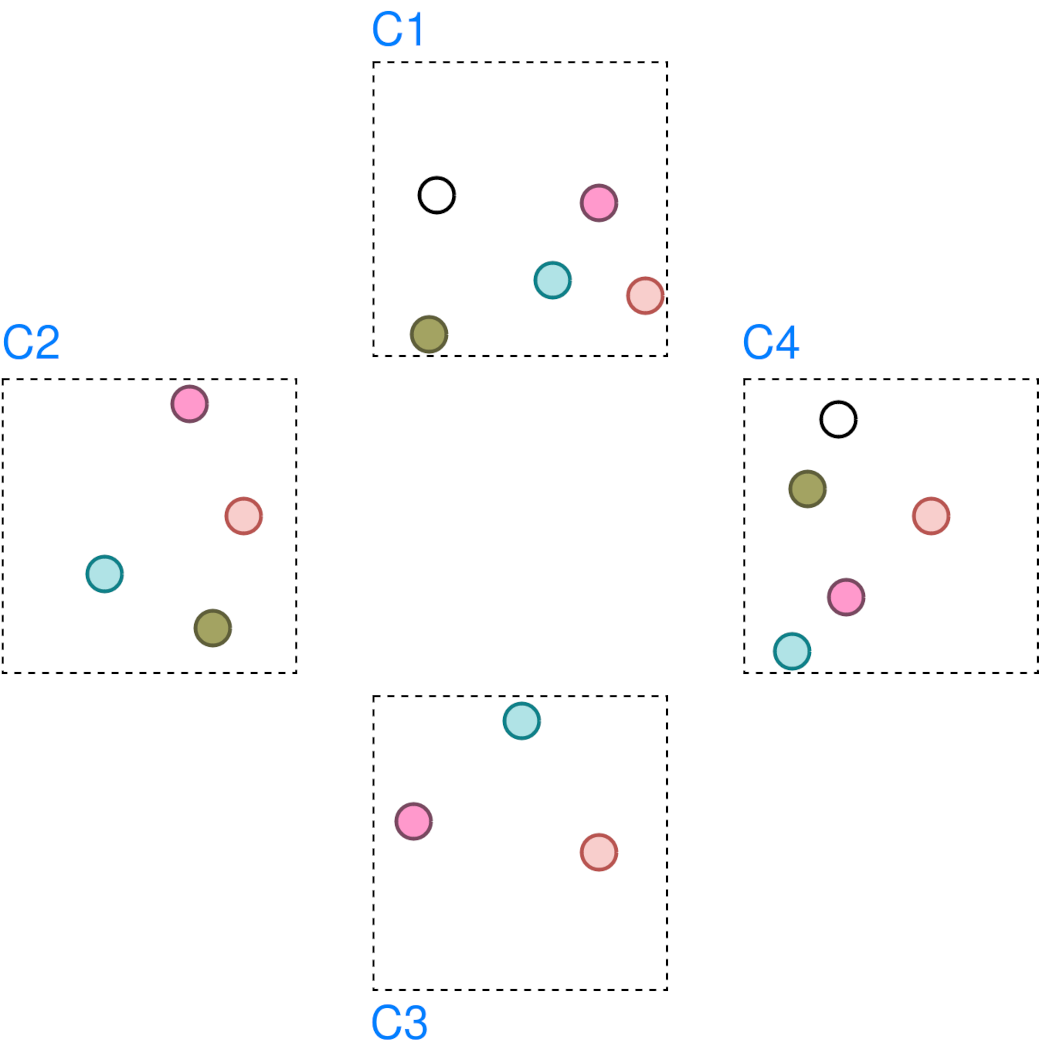}
    \vspace{0pt}
    \caption{Resulting clusters from Table~\ref{tb:pipeline}.}
    \vspace{20pt}
    \label{fig:clust-res}
\end{figure}

\subsection{Hierarchical Design} \label{hierarchy}

The whole pipeline explained in subsection \ref{clustering} can be repeated
on the final set of peak nodes as a new input (row $l_2$ in Table~\ref{tb:pipeline}).
Multi-level approach demands an aggregation of the features for each connected component
$q^{(l)}_i$ from the level $l$, which is replaced with a single node $v^{(l + 1)}_i$ on
the level $l + 1$. The node embeddings $h^{(l + 1)}_i$ of the next level is defined as a
concatenation of the peak node features $\tilde{h}^{l}_{q_{i}}$ and the mean node features
$\bar{h}^{l}_{q_{i}}$:
\begin{eqnarray}\label{eq:next_level1}
h^{(l + 1)}_i = [\tilde{h}^{l}_{q_{i}}, \bar{h}^{l}_{q_{i}}].
\end{eqnarray}
In the SGC-CCA approach, \textbf{we are passing the ground plane coordniates through
the levels} as well. Their aggregation for the $q^{(l)}_i$ is implemented as a mean value:
\begin{eqnarray}\label{eq:next_leve2}
(\hat{x}^{(l + 1)}_i, \hat{y}^{(l + 1)}_i)= (\frac{1}{Z}\sum_{j=1}^{Z}\hat{x}^{(l)}_j, \frac{1}{Z}\sum_{j=1}^{Z}\hat{y}^{(l)}_j).
\end{eqnarray}

After the aggregation step, repeating the \ref{clustering} provides the ability to merge
some of the connected components in order to reach the right level of granularity. The total
number of levels $L$ is a hyperparameter. The algorithm stops when the set $E'$ is empty
in the current level, or the level number $L$ is reached.
The last set of peaks defines the final cluster labels. Each peak label is propagated back
to the appropriate set of input nodes. Figure~\ref{fig:clust-res} depicts clusters
of the input instances, assigning the final peak color from the cell $(l_2, c)$
in Table~\ref{tb:pipeline}.

\subsection{Training}

\subsubsection{Ground-truth Graphs} \label{gt graphs}
The previously presented multi-level clustering algorithm assumed an inference time
scenario. The training procedure relies on the set of ground-truth graphs
$\{G^{(l)} \mid l \in [1, L]\}$ per each scene. They are created by skipping \ref{encoding}
\textit{Graph Encoding} and \ref{linkage prediction} \textit{Linkage Prediction and Node Density}
steps with next substitutions in the \ref{decoding} \textit{Graph Decoding}:
\begin{enumerate}
    \item \textit{Node density}: Instead of the estimation in Equation~\ref{eq:d_est},
    here we used the ground-truth node density defined in Equation~\ref{eq:d_gt}
    \begin{eqnarray}\label{eq:d_gt}
    d_i = \frac{1}{k}\sum_{j=1}^{k}(\mathds{1}(r_i = r_j) - \mathds{1}(r_i \neq r_j))a_{ij},
    \end{eqnarray}
    where $\mathds{1}$ is an indicator function and $r_i$ is a pedestrian label of the node $v_i$.  
    \item \textit{Edge filtering}: The subset of outgoing edges for the node $v_i$ is
    formed as:
    \begin{eqnarray}\label{eq:edge_subset_gt}
    \varepsilon(i) = \{j \mid (v_i, v_j) \in E \wedge d_i \leq d_j \wedge a_{ij} \geq p_{\tau}\}.
    \end{eqnarray}
    It can be noticed that ground-truth node density is used and the similarity (adjacency score)
    $a_{ij}$ is thresholded instead of predicted probability $\hat{r}_{ij}$ in
    Equation~\ref{eq:edge_subset}. Afterwards, the most similar node selection is performed
    again by adjacency score $a_{ij}$ unlike the Equation~\ref{eq:edge_filter}, wherein the
    edge coefficient is used:
    \begin{eqnarray}\label{eq:edge_filter_gt}
    j = \argmax_{k \in \varepsilon(i)} a_{ik}
    \end{eqnarray}
\end{enumerate}
\ref{hierarchy} \textit{Hierarchical Design} is applied in the way it is explained on top of the
modified \ref{decoding} \textit{Graph Decoding}.

Once the set of ground-truth graphs is acquired, the \ref{encoding} \textit{Graph Encoding}
is executed specifically for \textit{Linkage Prediction} and GCN training.
The predicted probabilities are then utilized for comparison against labeled connectivity,
thereby facilitating loss calculation.
\subsubsection{Loss}

Hi-LANDER \citep{xing2021learning} utilizes both the linkage and the node density prediction
in the composite loss implementation. It consists of connectivity and density loss.
\textbf{In our study, we determined that optimizing the connectivity loss alone was adequate for
our task.} This loss is specified as the binary cross-entropy loss per edge $l_{ij}$,
based on their linkage prediction $\hat{r}_{ij}$ (Equation~\ref{eq:conn_loss} and Equation~\ref{eq:bce}).
\begin{eqnarray}\label{eq:conn_loss}
L = -\frac{1}{|E|}\sum_{(v_i, v_j) \in E}^{}l_{ij}
\end{eqnarray}

\begin{eqnarray}\label{eq:bce}
l_{ij} = \mathds{1}(r_i = r_j)\log\hat{r}_{ij} + \mathds{1}(r_i \neq r_j)\log(1 - \hat{r}_{ij})
\end{eqnarray}
The indicator function $\mathds{1}(r_i = r_j)$ is the ground-truth label which indicates if two
nodes belong to the same cluster (the same pedestrian for this specific case).

Graphs from all levels of all scenes are considered equally and independently. They form a general
pool of ground-truth graphs, which are the partitioned in fixed-sized mini-batches for the
purpose of loss optimization.

\begin{table}[h]
\caption{Dataset combinations used for training (T) and validation (V).}
\vspace{10pt}
\centering
\begin{tabular}{c|ccc} 
\toprule
& Laboratory & Terrace & Basketball \\
\midrule
S1 & T & T & V \\
S2 & T & V & T \\
S3 & V & T & T \\
\bottomrule
\end{tabular}
\label{tb:subsets}
\end{table}


\section{Experiments}

\subsection {Dataset}

The EPFL video sequences dataset is
highly relevant for the described cross-camera association task.
It provides 4 overlapping Field of Views (FOVs), each containing a maximum of 9 pedestrians.
As authors of the GNN-CCA \citep{luna2022graph} explained, the EPFL dataset provides multiple
views of the same persons along with camera calibration parameters, facilitating global position
calculation. Thus, the EPFL is a very suitable for the approaches such as
GNN-CCA \citep{luna2022graph} and our proposed SGC-CCA. In the main experiment section, we kept the
sequences selection and combination strategy employed by the authors of GNN-CCA \citep{luna2022graph}.
This involves utilizing three distinct environments with different sets of pedestrians: Terrace - the outdoor
setup, Laboratory - the indoor setup, and Basketball - the sports court setup. Scene samples with
marked pedestrians across views can be seen in Figure~\ref{fig:seqs_preview}.

The training and validation (inference) datasets are constructed by combining three subsets
in various configurations. These combinations are outlined in Table~\ref{tb:subsets} and referred to as S1, S2, and S3.

\subsection {Implementation Details}

\subsubsection{Appearance Embeddings}
In these methodologies, pedestrian cropped images are encoded using pretrained ReID models.
\citet{luna2022graph} conducted a study on available relevant ReID feature extractors, 
concluding that model used in the \citep{braso2020learning} is a good compromise between
performance and the input size. This ResNet50 model is pretrained on the
Market1501 \citep{zheng2015scalable}, CUHK03 \citep{li2014deepreid}, and DukeMTMC \citep{ristani2016performance}.
It has an input size of 128x64 pixels and generates embedding output of size 256.

\subsubsection{Architecture Choices}

\paragraph{GNN} The utilized GNN architecture is a broadly used vanilla GCN \citep{kipf2017semisupervised}
model. We also explored the GAT \citep{velivckovic2017graph} alternative, but it didn't show
a significant difference. The input node feature size is the visual appearance feature extractor's
output size of 256. Our baseline experiment is conducted with 2 message passing steps where the
resulting node embedding size is 48.

\paragraph{Edge Encoder} The edge encoder $\theta$ in Equation~\ref{eq:edge_cls} is an MLP
consisting of 2 Linear layers with PReLU \citep{he2015delving} activation and a final sigmoid regressor.

\paragraph{Levels} The hierarchical framework is designed using $L = 3$ levels, with the
early-stopping paradigm. It relies on checking if any new edge is added in the current level
or if none of the peaks could be additionally connected. If so, the pipeline can be stopped
on the current level, and the current set of peaks defines detected clusters.

\paragraph{Batch} As discussed in the \ref{gt graphs} \textit{Ground-truth Graphs}, ground-truth
graphs are created for each level of each scene's affinity graph. They form a single pool of
graphs that are batched for training purposes.

\subsubsection{Training Procedure}

The training is performed using the batch size 48 for 200 epochs. The GCN and MLP models are
optimized using the Adam \citep{kingma2014adam} method and one-cycle cosine learning rate
scheduler \cite{He2018BagOT} with a base learning rate of 0.07. The regularization of the GCN
model is performed by the dropout technique of value 0.1. The edge probability threshold $p_{\tau}$
is assigned a value of 0.2.

\subsection{Evaluation Metrics}

Given that the task of data association in this study boils down to clustering person detections
across various views, we opt to utilize clustering performance metrics as employed in
GNN-CCA \citep{luna2022graph}.

\paragraph{Adjusted Rand Index (ARI)} This metric \citep{hubert1985comparing} is the similarity
measure between two clusterings. It is directly related to the number of pairs of instances
sharing the same or different labels in both the predicted and true clusterings. The term
"Adjusted" indicates that it is adjusted for chance, implying that its value approaches 0 for
random labeling and 1 when the clustering is optimal.

\paragraph{Adjusted Mutual Information (AMI)} AMI \citep{vinh2009information} reflects the
mutual information of two assignments - predicted and ground-truth cluster labels, ignoring
adopted label values. Even if we permute label values among predicted or ground-truth clusters,
AMI remains the same. That is a measure of agreement between two assignments. Like ARI,
this metric is also normalized against chance, to avoid high value for randomly assigned cluster labels.

\paragraph{V-measure (V-m)} This metric \citep{rosenberg2007v} is defined as a harmonic mean
between Homogeneity (\textbf{H}) and Completeness score (\textbf{C}). Homogeneity is the measure of the ground-truth labels
diversity among data points inside of the predicted cluster. If each cluster contains only items
of one identity, then Homogeneity is satisfied. Completeness score reflects the level of assigning
all members of one identity to the same cluster. As aforementioned metrics, these scores are also
independent of the absolute values of the labels.\\

All described metrics have a value of 1 for the perfect labeling (in our reports it is scaled to 100).

\begin{table}[h]
\caption{Comparison of different approaches, using pretrained ResNet50 model on Market-1051,
CUHK03 and DukeMTMC as feature extractor.}
\vspace{10pt}
\centering
\begin{tabular}{c|cccccc} 
\toprule
& \textbf{Method} & \textbf{ARI} & \textbf{AMI} & \textbf{H} & \textbf{C} & \textbf{V-m} \\
\midrule
S1 && 51.65 & 56.94 & 85.55 & 74.41 & 79.23 \\
S2 & Geom. approach & 44.17 & 50.62 & 78.79 & 71.33 & 73.04 \\
S3 & + Appearance \citep{lima2021generalizable} & 73.25 & 75.27 & 94.40 & 82.33 & 87.31 \\
avg. && \underline{56.36} & \underline{60.94} & \underline{86.25} & \underline{76.02} & \underline{79.86} \\
\midrule
S1 && 82.99 & 85.12 & 94.23 & 89.97 & 91.94 \\
S2 & GNN-CCA \citep{luna2022graph} & 83.07 & 86.77 & 93.59 & 92.03 & 92.66 \\
S3 & (reported) & 88.24 & 91.07 & 93.63 & 95.59 & 94.42 \\
avg. && \underline{84.76} & \underline{87.65} & \underline{93.81} & \underline{92.53} & \underline{93.00} \\
\midrule
S1 && 79.70 & 82.59 & 92.48 & 88.98 & 90.52 \\
S2 & GNN-CCA \citep{luna2022graph} & 82.17 & 85.61 & 94.44 & 91.12 & 92.55 \\
S3 & (reproduced) & 83.51 & 85.16 & 96.54 & 87.49 & 89.40 \\
avg. && \underline{81.79} & \underline{84.45} & \underline{94.49} & \underline{89.20} & \underline{90.82} \\
\midrule
S1 && 85.69 & 88.49 & 93.60 & 93.04 & 93.17 \\
S2 & SGC-CCA & 88.05 & 90.46 & 94.52 & 94.43 & 94.42 \\
S3 & (ours) & 90.42 & 92.28 & 95.69 & 95.14 & 95.33 \\
avg. && \underline{\textbf{88.05}} & \underline{\textbf{90.41}} & \underline{\textbf{94.60}} & \underline{\textbf{94.20}} & \underline{\textbf{94.31}} \\
\bottomrule
\end{tabular}
\label{tb:results}
\end{table}

\begin{figure}[h]
\centering
\includegraphics{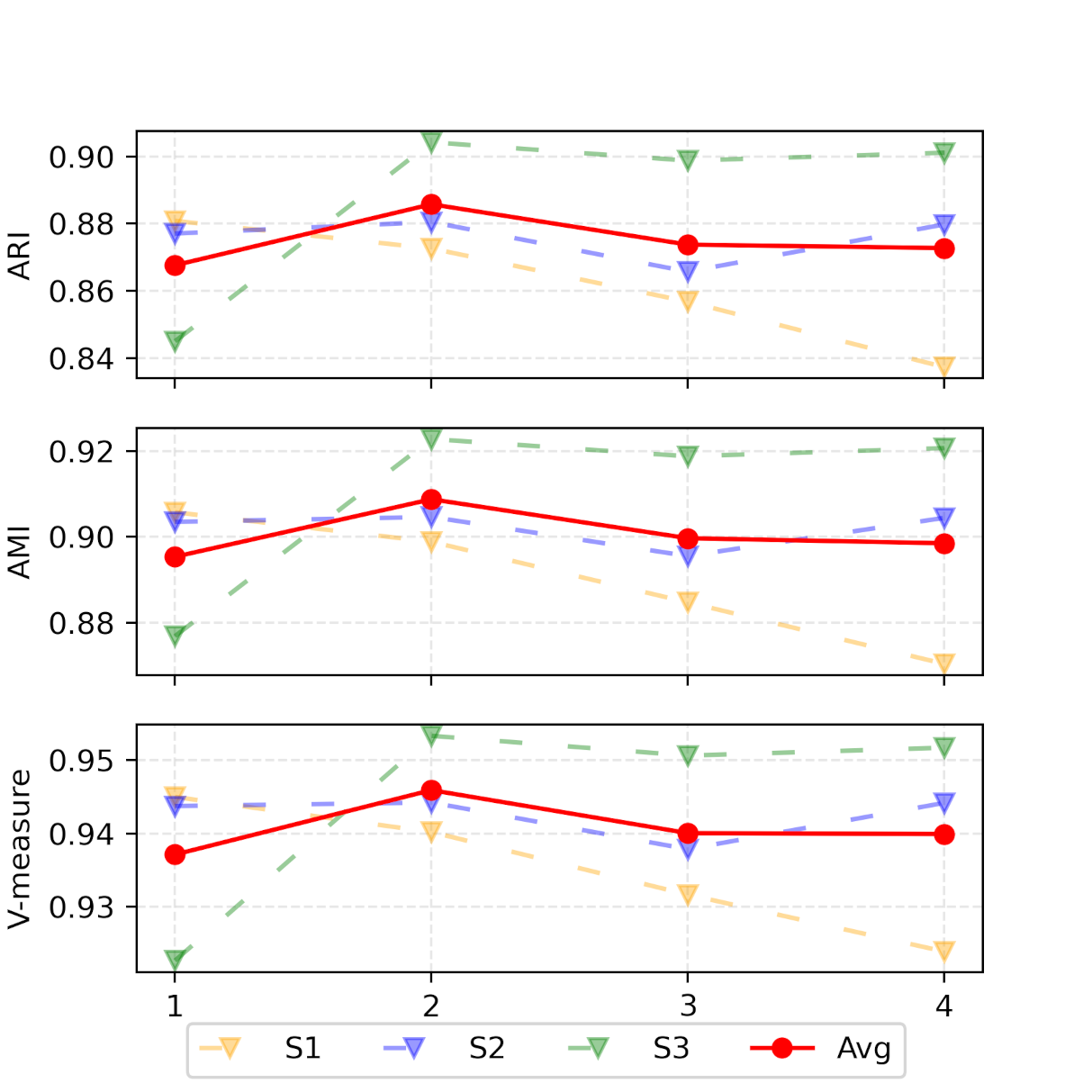}
\vspace{0pt}
\caption{Metrics for different number of message passing steps.}
\vspace{20pt}
\label{fig:mpsteps-results}
\end{figure}

\subsection {Results} \label{results}
Table~\ref{tb:results} presents a comparison of the proposed SGC-CCA method with previously
explored methods in \citep{luna2022graph}. We measureed ARI, AMI, Homogeneity, Completeness, and V-measure
for each dataset configuration S1-3 and aggregated them as the average value. As relevant comparable
methods, we consider the GNN-CCA \citep{luna2022graph} and the combination of geometrical and appearance features approach
\citep{lima2021generalizable}, which appears as the closest opponent to the method proposed
in \citep{luna2022graph}. Because only the S1 checkpoint of the GNN-CCA \citep{luna2022graph} is available online,
we tried to reproduce the model training for all S1-3 setups. We used the available hyperparameter values
in this study. For those that were not provided, we retained the default values from the code repository.
The reproduced results are reported in the separate row of the Table~\ref{tb:results}.
However, SGC-CCA shows the best performance on average across all clustering metrics.

We studied the impact of different message passing steps. The saturation of all metrics is achieved
for the value of 2, as can be noticed in Figure~\ref{fig:mpsteps-results}.

\begin{table}[h]
\caption{Clustering metrics of SGC-CCA over all training/validation/test dataset setups.}
\vspace{10pt}
\centering
\begin{tabular}{c|ccccc}
\toprule
\textbf{Setup} & \textbf{ARI} & \textbf{AMI} & \textbf{H} & \textbf{C} & \textbf{V-m} \\
\midrule
BLT & 85.33 & 88.43 & 93.72 & 92.66 & 93.05 \\
BTL & 85.84 & 88.65 & 93.84 & 92.19 & 92.85 \\
LBT & 87.11 & 89.86 & 94.39 & 93.82 & 94.03 \\
LTB & 81.78 & 85.86 & 92.77 & 91.00 & 91.69 \\
TBL & 89.86 & 91.93 & 95.57 & 94.80 & 95.07 \\
TLB & 84.86 & 88.02 & 93.62 & 92.44 & 92.86 \\
\midrule
avg. & \textbf{85.80} & \textbf{88.79} & \textbf{93.98} & \textbf{92.82} & \textbf{93.26} \\
\bottomrule
\end{tabular}
\label{tb:generalization}
\end{table}

\subsubsection{Generalization}

We conducted separate sets of experiments to assess the generalization capabilities
of the proposal. Although the main results section \ref{results} relies on the training
and validation setup from the independent domains, we aim to show that high performance
is kept even after the model selection during the training,  which is guided by
the validation dataset. Once the training is finished, we evaluate the model on the third independent set.
Given our three distinct environments, the training is performed over all six
permutations of \emph{training/validation/test} configurations. Table~\ref{tb:generalization}
contains particular results for each setup, along with the final averaged metrics in the last row.
The column \textbf{Setup} contains encoded names of the dataset combinations, where, for instance,
\emph{BTL} denotes Basketball/Laboratory/Terrace sequences as the \emph{training/validation/test}
datasets. These results indicate strong generalization capabilities of the proposed
method.


\section{Conclusions}

This work shows that the carefully tailored, general-purpose, supervised clustering
method Hi-LANDER \citep{xing2021learning} outperforms the state-of-the-art GNN
cross-camera data association method GNN-CCA \citep{luna2022graph} across all
clustering metrics. Furthermore, this approach avoids any graph post-processing
such as \emph{affinity graph splitting} or \emph{edge pruning}. Consequently,
these heuristics are left to be inferred from the training set, facilitated
by the sophisticated hierarchical architecture of \citep{xing2021learning}.

Moreover, the generalization experiments indicate that our proposal maintains
the same level of quality even when applied to entirely different environments
compared to the training and validation dataset.



\bibliography{mybibfile}

\begin{thebibliography}{29}
\providecommand{\natexlab}[1]{#1}
\providecommand{\url}[1]{\texttt{#1}}
\expandafter\ifx\csname urlstyle\endcsname\relax
  \providecommand{\doi}[1]{doi: #1}\else
  \providecommand{\doi}{doi: \begingroup \urlstyle{rm}\Url}\fi

\bibitem[Bras{\'o} and Leal-Taix{\'e}(2020)]{braso2020learning}
G.~Bras{\'o} and L.~Leal-Taix{\'e}.
\newblock Learning a neural solver for multiple object tracking.
\newblock In \emph{Proceedings of the IEEE/CVF conference on computer vision
  and pattern recognition}, pages 6247--6257, 2020.

\bibitem[Ester et~al.(1996)Ester, Kriegel, Sander, Xu,
  et~al.]{ester1996density}
M.~Ester, H.-P. Kriegel, J.~Sander, X.~Xu, et~al.
\newblock A density-based algorithm for discovering clusters in large spatial
  databases with noise.
\newblock In \emph{kdd}, volume~96, pages 226--231, 1996.

\bibitem[Gilmer et~al.(2017)Gilmer, Schoenholz, Riley, Vinyals, and
  Dahl]{gilmer2017neural}
J.~Gilmer, S.~S. Schoenholz, P.~F. Riley, O.~Vinyals, and G.~E. Dahl.
\newblock Neural message passing for quantum chemistry, 2017.

\bibitem[He et~al.(2015)He, Zhang, Ren, and Sun]{he2015delving}
K.~He, X.~Zhang, S.~Ren, and J.~Sun.
\newblock Delving deep into rectifiers: Surpassing human-level performance on
  imagenet classification.
\newblock In \emph{Proceedings of the IEEE international conference on computer
  vision}, pages 1026--1034, 2015.

\bibitem[He et~al.(2018)He, Zhang, Zhang, Zhang, Xie, and Li]{He2018BagOT}
T.~He, Z.~Zhang, H.~Zhang, Z.~Zhang, J.~Xie, and M.~Li.
\newblock Bag of tricks for image classification with convolutional neural
  networks.
\newblock \emph{2019 IEEE/CVF Conference on Computer Vision and Pattern
  Recognition (CVPR)}, pages 558--567, 2018.

\bibitem[Huang and Huang(2022)]{huang2022bevdet4d}
J.~Huang and G.~Huang.
\newblock Bevdet4d: Exploit temporal cues in multi-camera 3d object detection,
  2022.

\bibitem[Huang et~al.(2022)Huang, Huang, Zhu, Ye, and Du]{huang2022bevdet}
J.~Huang, G.~Huang, Z.~Zhu, Y.~Ye, and D.~Du.
\newblock Bevdet: High-performance multi-camera 3d object detection in
  bird-eye-view, 2022.

\bibitem[Hubert and Arabie(1985)]{hubert1985comparing}
L.~Hubert and P.~Arabie.
\newblock Comparing partitions.
\newblock \emph{Journal of classification}, 2:\penalty0 193--218, 1985.

\bibitem[Kingma and Ba(2014)]{kingma2014adam}
D.~P. Kingma and J.~Ba.
\newblock Adam: A method for stochastic optimization.
\newblock \emph{arXiv preprint arXiv:1412.6980}, 2014.

\bibitem[Kipf and Welling(2017)]{kipf2017semisupervised}
T.~N. Kipf and M.~Welling.
\newblock Semi-supervised classification with graph convolutional networks,
  2017.

\bibitem[Li et~al.(2014)Li, Zhao, Xiao, and Wang]{li2014deepreid}
W.~Li, R.~Zhao, T.~Xiao, and X.~Wang.
\newblock Deepreid: Deep filter pairing neural network for person
  re-identification.
\newblock In \emph{Proceedings of the IEEE conference on computer vision and
  pattern recognition}, pages 152--159, 2014.

\bibitem[Lima et~al.(2021)Lima, Roberto, Figueiredo, Simoes, and
  Teichrieb]{lima2021generalizable}
J.~P. Lima, R.~Roberto, L.~Figueiredo, F.~Simoes, and V.~Teichrieb.
\newblock Generalizable multi-camera 3d pedestrian detection.
\newblock In \emph{Proceedings of the IEEE/CVF conference on computer vision
  and pattern recognition}, pages 1232--1240, 2021.

\bibitem[Liu et~al.(2022)Liu, Wang, Zhang, and Sun]{liu2022petr}
Y.~Liu, T.~Wang, X.~Zhang, and J.~Sun.
\newblock Petr: Position embedding transformation for multi-view 3d object
  detection.
\newblock In \emph{European Conference on Computer Vision}, pages 531--548.
  Springer, 2022.

\bibitem[Liu et~al.(2023)Liu, Yan, Jia, Li, Gao, Wang, and
  Zhang]{liu2023petrv2}
Y.~Liu, J.~Yan, F.~Jia, S.~Li, A.~Gao, T.~Wang, and X.~Zhang.
\newblock Petrv2: A unified framework for 3d perception from multi-camera
  images.
\newblock In \emph{Proceedings of the IEEE/CVF International Conference on
  Computer Vision}, pages 3262--3272, 2023.

\bibitem[Lloyd(1982)]{lloyd1982least}
S.~Lloyd.
\newblock Least squares quantization in pcm.
\newblock \emph{IEEE transactions on information theory}, 28\penalty0
  (2):\penalty0 129--137, 1982.

\bibitem[Luna et~al.(2022)Luna, SanMiguel, Mart{\'\i}nez, and
  Carballeira]{luna2022graph}
E.~Luna, J.~C. SanMiguel, J.~M. Mart{\'\i}nez, and P.~Carballeira.
\newblock Graph neural networks for cross-camera data association.
\newblock \emph{IEEE Transactions on Circuits and Systems for Video
  Technology}, 33\penalty0 (2):\penalty0 589--601, 2022.

\bibitem[Otto et~al.(2017)Otto, Wang, and Jain]{otto2017clustering}
C.~Otto, D.~Wang, and A.~K. Jain.
\newblock Clustering millions of faces by identity.
\newblock \emph{IEEE transactions on pattern analysis and machine
  intelligence}, 40\penalty0 (2):\penalty0 289--303, 2017.

\bibitem[Ristani et~al.(2016)Ristani, Solera, Zou, Cucchiara, and
  Tomasi]{ristani2016performance}
E.~Ristani, F.~Solera, R.~Zou, R.~Cucchiara, and C.~Tomasi.
\newblock Performance measures and a data set for multi-target, multi-camera
  tracking.
\newblock In \emph{European conference on computer vision}, pages 17--35.
  Springer, 2016.

\bibitem[Rosenberg and Hirschberg(2007)]{rosenberg2007v}
A.~Rosenberg and J.~Hirschberg.
\newblock V-measure: A conditional entropy-based external cluster evaluation
  measure.
\newblock In \emph{Proceedings of the 2007 joint conference on empirical
  methods in natural language processing and computational natural language
  learning (EMNLP-CoNLL)}, pages 410--420, 2007.

\bibitem[Shi and Malik(2000)]{shi2000normalized}
J.~Shi and J.~Malik.
\newblock Normalized cuts and image segmentation.
\newblock \emph{IEEE Transactions on pattern analysis and machine
  intelligence}, 22\penalty0 (8):\penalty0 888--905, 2000.

\bibitem[Veli{\v{c}}kovi{\'c} et~al.(2017)Veli{\v{c}}kovi{\'c}, Cucurull,
  Casanova, Romero, Lio, and Bengio]{velivckovic2017graph}
P.~Veli{\v{c}}kovi{\'c}, G.~Cucurull, A.~Casanova, A.~Romero, P.~Lio, and
  Y.~Bengio.
\newblock Graph attention networks.
\newblock \emph{arXiv preprint arXiv:1710.10903}, 2017.

\bibitem[Vinh et~al.(2009)Vinh, Epps, and Bailey]{vinh2009information}
N.~Vinh, J.~Epps, and J.~Bailey.
\newblock Information theoretic measures for clusterings comparison: Variants.
\newblock \emph{Properties, Normalization and Correction for Chance}, 18, 2009.

\bibitem[Wang et~al.(2021)Wang, Guizilini, Zhang, Wang, Zhao, and
  Solomon]{wang2021detr3d}
Y.~Wang, V.~Guizilini, T.~Zhang, Y.~Wang, H.~Zhao, and J.~Solomon.
\newblock Detr3d: 3d object detection from multi-view images via 3d-to-2d
  queries, 2021.

\bibitem[Wang et~al.(2019)Wang, Zheng, Li, and Wang]{wang2019linkage}
Z.~Wang, L.~Zheng, Y.~Li, and S.~Wang.
\newblock Linkage based face clustering via graph convolution network.
\newblock In \emph{Proceedings of the IEEE/CVF conference on computer vision
  and pattern recognition}, pages 1117--1125, 2019.

\bibitem[Xing et~al.(2021)Xing, He, Xiao, Wang, Xiong, Xia, Wipf, Zhang, and
  Soatto]{xing2021learning}
Y.~Xing, T.~He, T.~Xiao, Y.~Wang, Y.~Xiong, W.~Xia, D.~Wipf, Z.~Zhang, and
  S.~Soatto.
\newblock Learning hierarchical graph neural networks for image clustering.
\newblock In \emph{Proceedings of the IEEE/CVF International Conference on
  Computer Vision}, pages 3467--3477, 2021.

\bibitem[Yang et~al.(2019)Yang, Zhan, Chen, Yan, Loy, and
  Lin]{yang2019learning}
L.~Yang, X.~Zhan, D.~Chen, J.~Yan, C.~C. Loy, and D.~Lin.
\newblock Learning to cluster faces on an affinity graph.
\newblock In \emph{Proceedings of the IEEE/CVF conference on computer vision
  and pattern recognition}, pages 2298--2306, 2019.

\bibitem[Yang et~al.(2020)Yang, Chen, Zhan, Zhao, Loy, and
  Lin]{yang2020learning}
L.~Yang, D.~Chen, X.~Zhan, R.~Zhao, C.~C. Loy, and D.~Lin.
\newblock Learning to cluster faces via confidence and connectivity estimation.
\newblock In \emph{Proceedings of the IEEE/CVF conference on computer vision
  and pattern recognition}, pages 13369--13378, 2020.

\bibitem[Zheng et~al.(2015)Zheng, Shen, Tian, Wang, Wang, and
  Tian]{zheng2015scalable}
L.~Zheng, L.~Shen, L.~Tian, S.~Wang, J.~Wang, and Q.~Tian.
\newblock Scalable person re-identification: A benchmark.
\newblock In \emph{Proceedings of the IEEE international conference on computer
  vision}, pages 1116--1124, 2015.

\bibitem[Zhu et~al.(2011)Zhu, Wen, and Sun]{zhu2011rank}
C.~Zhu, F.~Wen, and J.~Sun.
\newblock A rank-order distance based clustering algorithm for face tagging.
\newblock In \emph{CVPR 2011}, pages 481--488. IEEE, 2011.

\end{thebibliography}

\end{document}